\documentclass[10pt,twocolumn,letterpaper]{article}

\usepackage[pagenumbers]{cvpr} %

\usepackage{times}
\usepackage{epsfig}

\usepackage[dvipsnames]{xcolor}

\usepackage{graphicx}
\usepackage{comment}
\usepackage{amsmath,amssymb} %
\usepackage{dsfont}
\usepackage{color, colortbl}
\usepackage{enumitem}

\usepackage{xcolor}
\usepackage{nicefrac}
\usepackage{booktabs}
\usepackage{placeins}
\usepackage{pifont}
\usepackage{subcaption}
\usepackage{xspace}
\usepackage{arydshln}
\usepackage{nicefrac}

\usepackage{algorithm}
\usepackage{algpseudocode}
\algrenewcommand\algorithmicindent{1.0em}%

\usepackage{array}
\usepackage{booktabs}
\usepackage{adjustbox}

\usepackage{listings}
\usepackage{multirow}

\usepackage[colorlinks,citecolor=red,urlcolor=blue,bookmarks=false,hypertexnames=true,pagebackref=true,breaklinks=true]{hyperref}

\renewcommand{\backref}[1]{}
\renewcommand{\backrefalt}[4]{%
  \ifcase #1 %
No citations.%
  \or
(page #2).%
  \else
(pages #2).%
  \fi%
}

\definecolor{BrickRed}{HTML}{B6321C}
\definecolor{RoyalBlue}{HTML}{0071BC}
\definecolor{PineGreen}{HTML}{008B72}
\definecolor{bluefig}{HTML}{5B9BD5}
\definecolor{Gray}{gray}{0.9}

\let\thetaold\theta
\renewcommand{\theta}{\boldsymbol{\thetaold}}

\DeclareMathOperator*{\argmax}{argmax} %
\newcommand{\parag}[1]{\vspace{0.2cm}\noindent\textbf{#1}.\ }

\newcolumntype{R}[2]{%
    >{\adjustbox{angle=#1,lap=\width-(#2)}\bgroup}%
    l%
    <{\egroup}%
}

\usepackage[capitalize]{cleveref}
\crefname{section}{Sec.}{Secs.}
\Crefname{section}{Section}{Sections}
\Crefname{table}{Table}{Tables}
\crefname{table}{Tab.}{Tabs.}

\begin{document}

\title{\vspace{-2em}CoMFormer: Continual Learning in Semantic and Panoptic Segmentation}

\author{
Fabio Cermelli\ $^{1,2}$\thanks{Work done during the visiting period at Sorbonne Université.}, Matthieu Cord\ $^{3,4}$, Arthur Douillard\ $^{3}$\thanks{Work done while at Heuritech, currently affiliated to DeepMind.}\\
$^1$Politecnico di Torino, $^2$Italian Institute of Technology, $^3$ Sorbonne Université, $^4$ valeo.ai\\
{\tt\small fabio.cermelli@polito.it, matthieu.cord@lip6.fr, ar.douillard@gmail.com } \\
}

\maketitle

\begin{abstract}
Continual learning for segmentation has recently seen increasing interest. However, all previous works focus on narrow semantic segmentation and disregard panoptic segmentation, an important task with  real-world impacts. %
In this paper, we present the first continual learning model capable of operating on both semantic and panoptic segmentation. 
Inspired by recent transformer approaches that consider segmentation as a mask-classification problem, we design CoMFormer.
Our method carefully exploits the properties of transformer architectures to learn new classes over time. Specifically, we propose a novel adaptive distillation loss along with a mask-based pseudo-labeling technique to effectively prevent forgetting.
To evaluate our approach, we introduce a novel continual panoptic segmentation benchmark on the challenging ADE20K dataset. 
Our CoMFormer outperforms all the existing baselines by forgetting less old classes but also learning more effectively new classes. In addition, we also report an extensive evaluation in the large-scale continual semantic segmentation scenario showing that CoMFormer also significantly outperforms state-of-the-art methods.
\end{abstract}

\section{Introduction}
\label{sec:intro}
Image segmentation is a fundamental computer vision problem that enables machines to assign an image's pixels to discrete segments. Multiple segmentation tasks have been defined depending on the segments definitions. Semantic segmentation clusters pixels by classes, merging in a single segment pixels belonging to instances of the same class. Panoptic segmentation assigns to every pixel a semantic class while separating different instances into different segments. This latter kind of segmentation has real-world impacts in autonomous robots and vehicles \cite{chen2021panoptic, neuhold2017mapillary}.

\begin{figure}
    \includegraphics[width=\linewidth]{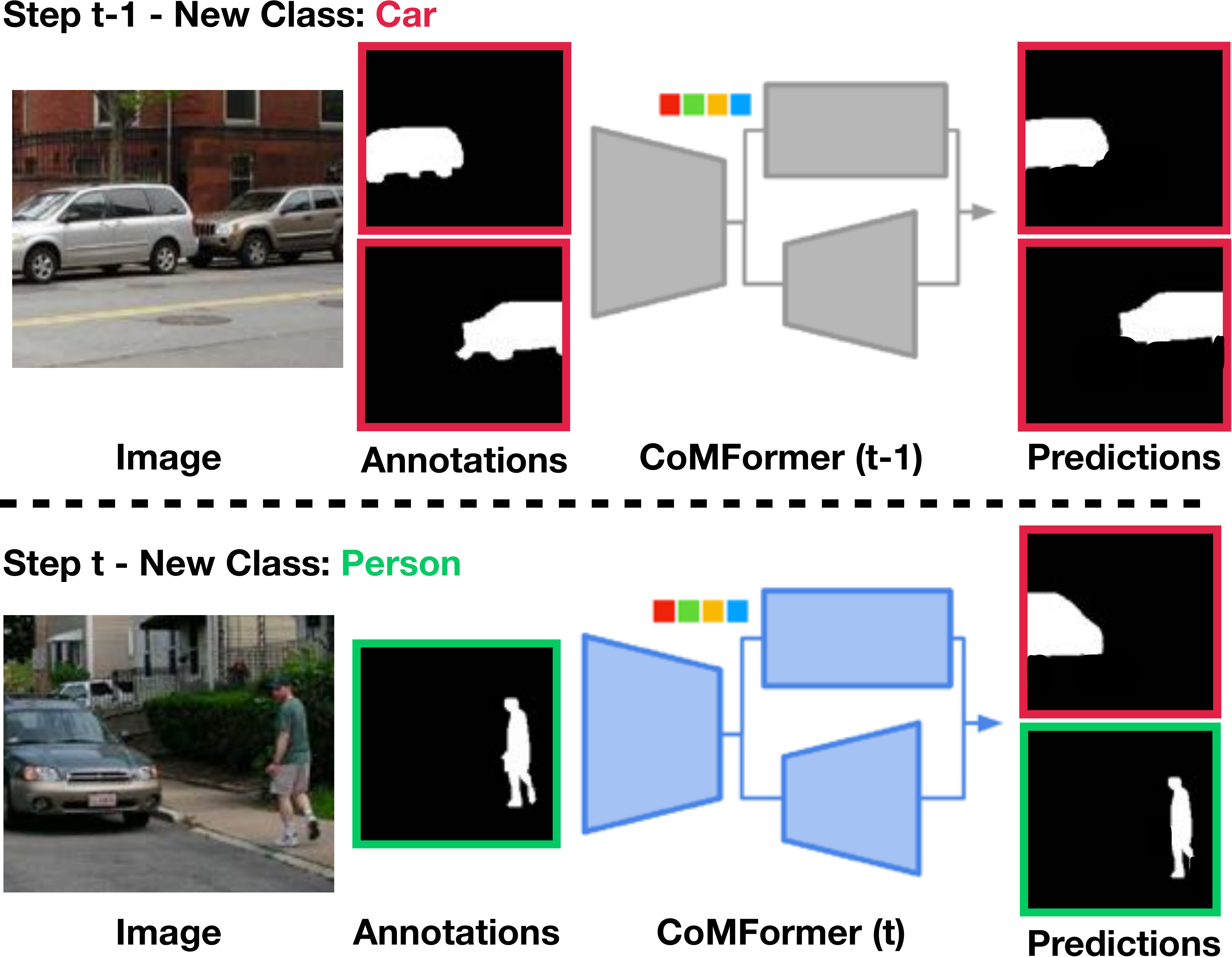}
    \caption{\textbf{Illustration of our model, CoMFormer, operating in continual segmentation}. Relying on the mask classification paradigm, it is able to cope with both continual semantic and panoptic segmentation without any modification by predicting masks for both old (\eg \textit{car} in red) and new (\eg \textit{person} in green) classes. The figure reports two classes and no “stuff" (\eg \textit{road}, \textit{building}) only for illustration purposes.}\vspace{-1em}
    \label{fig:intro_continual_panoptic}
\end{figure}

Despite tremendous progress in image segmentation, the current approaches are trained on a static dataset with a predefined set of classes. Whenever an update of the model is required to fit new classes, the common solution is to train a model from scratch on the union of the old and new class data. A computationally more efficient solution would be to fine-tune the existing model solely on the new class data. Unfortunately, this approach would cause a \textit{catastrophic forgetting} \cite{french1999catastrophicforgetting} of the old classes on which the model performance would be extremely degraded.

The problem of updating the knowledge of the model over time is typically referred as continual learning. It has been traditionally studied in the context of image classification \cite{li2018lwf, kirkpatrick2017ewc, rebuffi2017icarl, rusu2016progressive, douillard2020podnet, douillard2021dytox} and is gaining attention on the segmentation task \cite{michieli2019ilt, cermelli2020modelingthebackground, douillard2020plop, zhang2022rcil, cermelli2022modeling} due to the more realistic applications and the additional challenges it introduces, such as the background shift \cite{cermelli2020modelingthebackground}.
However, current state-of-the-art methods mainly focus on semantic segmentation and are not designed to work in other segmentation tasks, %
strongly limiting their application in the real world. 

In this paper, we design the first method operating in both continual semantic and panoptic segmentation, as illustrated in \autoref{fig:intro_continual_panoptic}.
Our method, CoMFormer (\textbf{Co}ntinual \textbf{M}ask\textbf{Former}), takes inspiration from recent %
transformer architectures \cite{cheng2021maskformer, cheng2021masked}, approaching segmentation as a \textit{mask classification} problem. Instead of predicting a class probability for each pixel, as in previous semantic segmentation works \cite{long2015fcn, chen2017deeplabv3}, it predicts a set of binary masks, each associated with a single class prediction, effectively addressing both segmentation tasks without any modification in the training architecture and procedure. 
Differently from previous works \cite{cheng2021maskformer, cheng2021masked}, however, CoMFormer forces the output binary masks to be mutually exclusive to one another: a pixel can only be predicted by a single binary mask to prevent having several masks classifying the same pixel with different classes. {This behavior is crucial in continual learning to reduce the interference among old and new classes}. %

Furthermore, CoMFormer introduces a novel adaptive distillation loss to alleviate forgetting. It enforces consistency of the model's classification predictions across learning steps %
only when it is useful to remember old classes, ensuring a better tradeoff between rigidity (not forgetting old classes) and plasticity (learning efficiently new classes).
Finally, since at each training iteration the dataset reports annotations only for the current classes, we design a mask-based pseudo-labeling technique to generate annotations for the old classes, effectively alleviating forgetting. To reduce the noise, we consider the prediction confidence and we avoid interference with ground-truth annotations. 

We validate CoMFormer on both continual segmentation tasks. For panoptic segmentation, we define a new benchmark relying on the challenging ADE20K where we demonstrate that CoMFormer largely outperforms all previous baselines. 
On semantic segmentation, we show that CoMFormer outperforms the existing state-of-the-art methods on every setting of the large-scale ADE20K benchmark. 

To sum up, the contributions of this paper are as follows:
\begin{itemize}[noitemsep,topsep=0pt]
\item We introduce continual panoptic segmentation which has real-world impacts in addition to being significantly more challenging than previous benchmarks.
\item We propose CoMFormer to tackle both continual panoptic and semantic segmentation. To avoid forgetting, we design a novel adaptive distillation and an efficient mask-based pseudo-labeling strategy.
\item Through extensive quantitative and qualitative benchmarks, we showcase the state-of-the-art performance of our model on both continual segmentation tasks.
\end{itemize}
\vspace{-0.5em}

\begin{figure*}[t]
    \includegraphics[width=\textwidth]{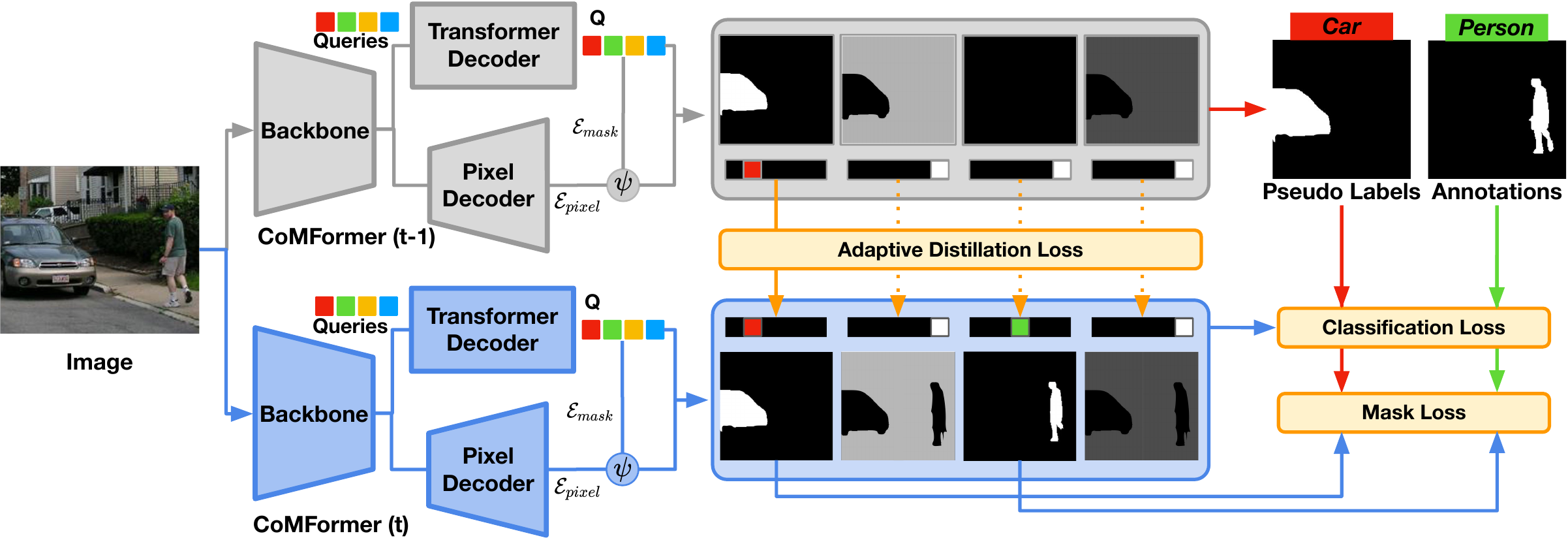}
    \caption{\textbf{Overview of CoMFormer}. At learning step t, both the current (bottom) and old (top) models output, for each input query, a pair made of a class probability distribution and a binary mask. While learning the new class (\textit{person}), we regularize the current model to prevent forgetting the old classes (\textit{car}) by: (i) generating pseudo-labels for the old classes present in the image and (ii) by applying an adaptive distillation loss that reweights the contribution of each output based on the information it carries on the old classes. $\psi$ indicates the dot product between the mask embeddings $\mathcal{E}_{mask}$ and the pixel embeddings $\mathcal{E}_{pixel}$ followed by the softmax activation.
    }\label{fig:model} \vspace{-1em}
\end{figure*}

\section{Related Works}
\label{sec:related}
\vspace{-1.5em}\noindent\paragraph{Semantic and Panoptic Segmentation.}
The two tasks have been traditionally treated separately, with specialized architectures proposed for either one or the other task, without interoperability. 
Semantic segmentation has been traditionally addressed as a per-pixel classification task. Fully-convolutional network \cite{long2015fcn} dominated the field by aggregating long-range dependencies in the features map \cite{chen2018deeplab, chen2017deeplabv3, zhao2017PSPNet} and exploiting contextual information \cite{fu2019DANet, huang2020ccnet, yuan2018ocnet, zhao2018psanet, yuan2020ocr}. Recently, transformers \cite{jain2021semask, xie2021segformer, strudel2021segmenter} are replacing convolutions by integrating long-range dependencies at every layer.
Panoptic segmentation \cite{kirillov2019panoptic} has been proposed to unify semantic and instance segmentations. Initially, methods proposed to combine each task-specific architectures \cite{cheng2020panoptic, kirillov2019panopticpyr, li2022fully, Porzi_2019_CVPR} or defined new specialized architectures and objective functions for the panoptic task \cite{carion2020end, wang2021max, wang2020axial}, drifting from a general solution for all the segmentation tasks.
Recently, methods addressing segmentation as a mask classsification problem \cite{wang2021max, yu2022kMax, cheng2021masked, cheng2021maskformer} have been proposed, introducing a transformer architecture able to solve multiple tasks at once. MaskFormer \cite{cheng2021maskformer} was the first to propose a single architecture to address both panoptic and semantic segmentation. Mask2Former \cite{cheng2021masked} improves it by adopting multi-scale features, masked attention and optimization tricks. Concurrently, kMaX-DeepLab \cite{yu2022kMax} extended \cite{wang2021max} proposing to reformulate the cross-attention as a clustering process.
Despite their effectiveness in standard training setting, these methods suffer from catastrophic forgetting. In this work, we aim to extend their capability and we propose CoMFormer to learn to segment new classes over time.

\parag{Continual Segmentation} 
Continual learning is a long-standing field addressing the problem of learning new knowledge over time while avoiding catastrophic forgetting \cite{french1999catastrophicforgetting, thrun1998lifelonglearning, robins1995catastrophicforgetting}. Traditionally, it has been studied in the context of image classification \cite{rebuffi2017icarl, li2018lwf, kirkpatrick2017ewc, rusu2016progressive, douillard2020podnet, chaudhry2019AGEM, yan2021der, douillard2021dytox} %
but there is a growing interest in its applications in segmentation \cite{michieli2019ilt, cermelli2020modelingthebackground, cermelli2020fewshotcontinualsegm, douillard2020plop, cermelli2022incremental, douillard2021objectrehearsal, michieli2021continual, maracani2021recall, zhang2022rcil, cermelli2022modeling}. Continual semantic segmentation introduces additional challenges, as pointed out by \cite{cermelli2020modelingthebackground}. In particular, considering the task as a per-pixel classification problem, catastrophic forgetting is exacerbated by the background shift, where old classes pixels are treated as background in following training steps. \cite{cermelli2020modelingthebackground} proposed a revision of the standard knowledge distillation framework to deal with it. \cite{douillard2020podnet, michieli2019ilt} introduced a distillation loss preserving the representation at feature level while \cite{maracani2021recall} proposed an approach using old classes' examples to alleviate forgetting. Recently, \cite{zhang2022rcil} presented a technique that decouples learning the representation of the new classes while freezing the representation for the old ones. %
Differently from previous work, we extend the benchmarks to solve both semantic and the more challenging panoptic segmentation task. %

\parag{Continual Learning with Transformers}
Recent progresses made using tranformers for computer vision \cite{dosovitskiy2020vit, touvron2021cait, touvron2021deit, liu2021swin} attracted the attention of the continual learning community \cite{douillard2021dytox, wang2022learning, wang2022dualprompt}. In particular, DyTox \cite{douillard2021dytox} proposed %
to specialize the architecture on each task using a different task-specific token. Learning-to-Prompt \cite{wang2022learning} stores a pool of prompts that are employed to condition the whole forward execution of the patch tokens. To the best of our knowledge, transformers have been evaluated only in continual learning for image classification \cite{douillard2021dytox, li2021continualtransformer} and we are the first to use them in the continual segmentation task.

\section{CoMFormer}
\label{sec:method}
\subsection{Problem Definition} \label{sec:setting}
The goal of image segmentation is to learn a model able to (i) partition the image into a set of N regions represented by binary masks and (ii) produce a class-probability distribution associated with each region. The differences among segmentation tasks rely on the semantics of the masks: semantic segmentation groups all the pixels of a class while panoptic segmentation distinguishes different object instances. In the following, we provide a general formulation for the two tasks since they only differ in the construction of the binary masks in the dataset.

Continual segmentation aims to train the model in multiple learning steps $t = 1, \dots, T$, introducing at every step a new set of classes.
Formally, during the learning step $t$ a dataset $D_t$ consisting of a collection of image and label pairs is provided.
The label of each image takes the form of a set of $N^{gt}$ ground truth segments $z^{gt}\{(c_i^{gt}, m_i^{gt})|c_i^{gt} \in \mathcal{K}^t, m_i^{gt} \in \{0,1\}^{H \times W} \}_{i=1}^{N^{gt}}$ where $c_i^{gt}$ is the ground-truth class and $m_i^{gt}$ is binary mask $m^{gt}$, $\mathcal{K}^t$ is the set of classes introduced at step $t$, and $H$, $W$ are the height and width of the images. The goal of training step $t$ is to learn a model able to predict segments for all the seen classes $\mathcal{K}^{1:t}$.

We note that the dataset only contains annotations for the new classes $\mathcal{K}^t$ while not reporting ground-truth segments for old $\mathcal{K}^{1:t-1}$ or future $\mathcal{K}^{t+1:T}$ classes. 
Moreover, differently from previous works \cite{cermelli2020modelingthebackground}, old and future class segment annotations are simply absent during training rather than being collapsed into an artificial background class. %

\subsection{CoMFormer Architecture}
To solve semantic and panoptic segmentation tasks within a single method, we take inspiration from MaskFormer architectures \cite{cheng2021maskformer, cheng2021masked} considering segmentation a mask classification problem. It consists in predicting for each image a set of $N$ pairs made of a class prediction and a binary mask.
Formally, the output of a CoMFormer is $z = \{(p_i, m_i)\}_{i=1}^{N}$, where $p_i$ represents a class probability distribution over the seen classes, \ie $p_i \in \Delta^{K^t+1}$ with $K^t = |\mathcal{K}^{1:t}|$, and $m_i$ is a binary mask such that $m_i \in [0,1]^{H \times W}$. We note that $p_i$ contains an additional “no object" class (denoted $\varnothing$) to indicate that the mask does not correspond to any of the known $\mathcal{K}^{1:t}$ categories.

To obtain such output, the CoMFormer architecture, showcased in \autoref{fig:model}, is made of three components: (i) a backbone that extracts feature embeddings $\mathcal{E}_{feat}$, (ii) a transformer decoder that takes as input N learnable queries and $\mathcal{E}_{feat}$ to output $N$ per-segment embeddings $Q$, (iii) a pixel decoder that takes $\mathcal{E}_{feat}$ as input and extracts per-pixel embeddings $\mathcal{E}_{pixel}$. The output class probabilities $\{p_i \in \Delta^{K^t+1}\}_{i=1}^N$ are obtained by applying a linear classifier on $Q$, which is enlarged as new classes are presented. We obtain the mask predictions $m_i$ combining $Q$ and $\mathcal{E}_{pixel}$: first $Q$ is passed into a 2-layer MLP to obtain N mask embeddings $\mathcal{E}_{mask}$ and we then compute the dot product between the $i$-th mask embedding and the pixel embeddings, followed by the softmax activation: $m_i[h,w] = \text{softmax}(\mathcal{E}_{mask}[:,i]^T \cdot \mathcal{E}_{pixel}[:,h,w])$.

\begin{figure}[t]
    \includegraphics[width=\linewidth]{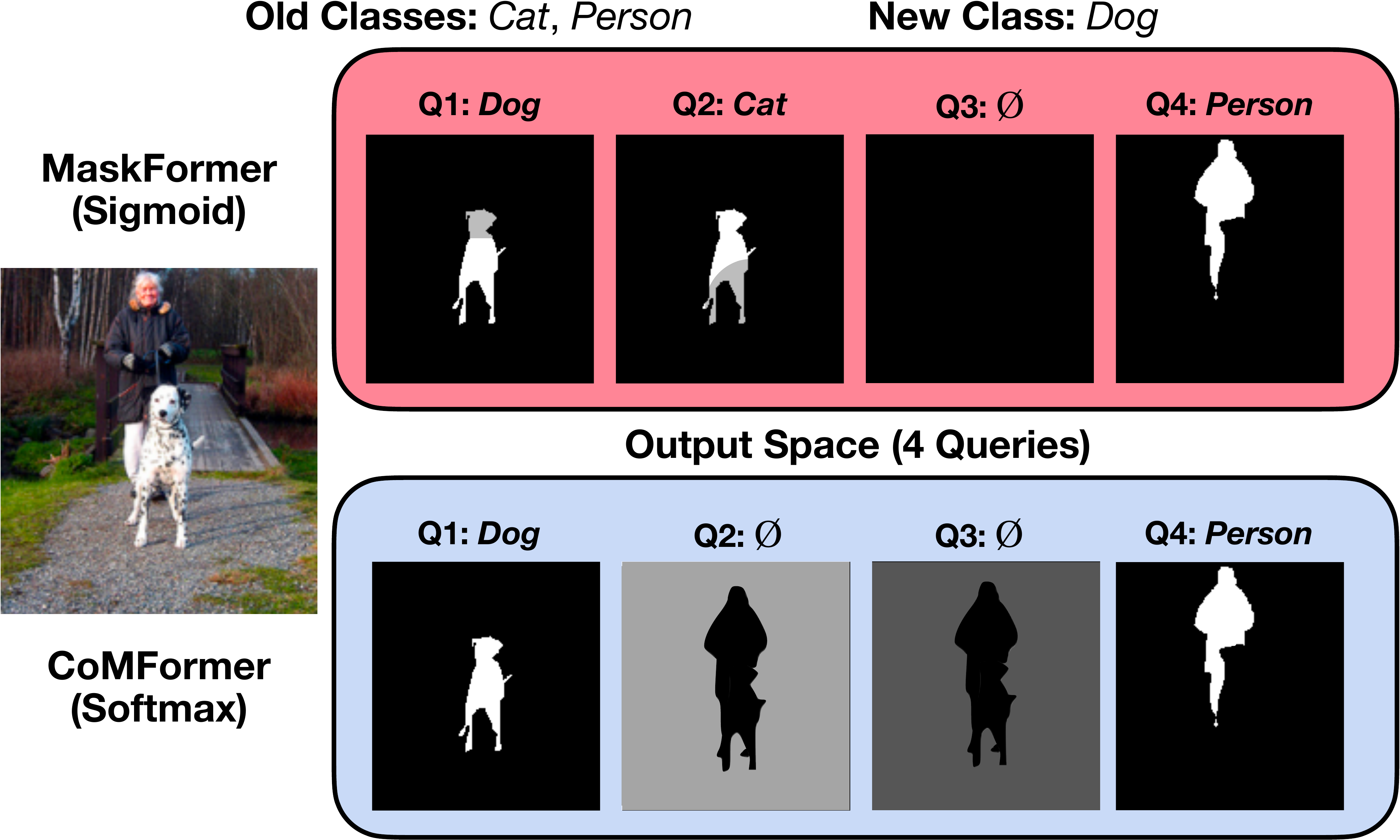}
    \caption{\textbf{Illustration of the mask predictions between sigmoid and softmax}. MaskFormer has overlap across masks (\eg in Query 1 and 2) predicted either as old or new classes, while CoMFormer avoids the issue using the softmax activation.}
    \label{fig:softmask} \vspace{-1em}
\end{figure}

We point out that employing the softmax contrasts sharply with MaskFormers \cite{cheng2021maskformer, cheng2021masked}, which use sigmoid activation to get the mask predictions. Using the sigmoid activation would result in the overlap of many masks on a single pixel, each potentially belonging to a distinct class, as seen in \cref{fig:softmask}. This is particularly important in the context of continual learning, when the model may predict two similar masks associated with two distinct classes (\eg \textit{cat} and \textit{dog} classes in \cref{fig:softmask}) thereby generating interference among old and new classes and degrading the model's performance.

\subsection{Learning without Forgetting} \label{sec:lwf}
To learn new classes, a simple solution is to fine-tune the model on the new dataset. However, because it lacks annotations for old classes, this operation results in catastrophic forgetting \cite{french1999catastrophicforgetting}. To alleviate the issue, we introduce two techniques for avoiding forgetting: an adaptive distillation loss and a mask-based pseudo-labeling strategy.

\parag{Adaptive Distillation Loss}
A common strategy to mitigate forgetting adopted by continual semantic segmentation approaches \cite{cermelli2020modelingthebackground, michieli2019ilt, douillard2020plop} is to distill the knowledge coming from the old model in the new one, either in form of features \cite{douillard2020plop, michieli2019ilt} or as classification probabilities \cite{cermelli2020modelingthebackground}.
We take inspiration from the latter solution and we design a novel distillation loss tailored to the CoMFormer architecture. %
In particular, analyzing the two components constituting the output space, we find the mask prediction being robust to forgetting while the classification part is heavily affected (see \cref{sec:ablation}). For this reason, we focus on the output class probabilities to design our distillation loss.

An adaptation of standard knowledge distillation works \cite{cermelli2020modelingthebackground} would force the new model to mimic the old model probabilities over all the N outputs, independently from their content. Formally, given the set of N probability distributions $\{p_i^{o}\}_{i=1}^{N}$ coming from the old model, a standard distillation loss is computed as
\begin{equation} \label{eq:ukd}
    \mathcal{L}_{KD} = - \frac{1}{N} \sum_{i=1}^{N} \sum_{k=1}^{K^{t-1}+1} p_i^o(k) \log \frac{\tilde{p}_i(k)}{p_i^o(k)},
\end{equation}
where $K^{t-1}+1$ is the $\varnothing$ class and $\tilde{p}_i(k)$ represents the unbiased probability distribution of $i$-th output for class $k$:
\begin{equation}
    \tilde{p}_i(k) =
    \begin{cases}
        \sum_{j \in \mathcal{K}^t} p_i(j) + p_i(\varnothing) & \text{if $k = \varnothing$},\\
        p_i(k) & \text{otherwise}.\\
    \end{cases}
\end{equation}
Note that, similarly to \cite{cermelli2020modelingthebackground}, it forces the sum of the model new class ($\mathcal{K}^t$) and the “no object" class ($\varnothing$) probabilities to be similar to the old model $\varnothing$ probability.

However, not every output probability brings relevant information: most of them are predicted as “no object" ($\varnothing$) and thus they do not carry enough details to remember the old classes while reducing the relative importance of the other outputs probabilities. 
For this reason, we re-weight their contribution based on the probability of not being $\varnothing$. Formally, we define our adaptive distillation loss as
\begin{equation}  \label{eq:qd}
    \mathcal{L}_{AD}(p, p^o) = - \sum_{i=1}^{N} \frac{\omega_i}{\sum_{i=1}^{N} \omega_i} \sum_{k=1}^{K^{t-1}+1} p_i^o(k) \log \frac{\tilde{p}_i(k)}{p_i^o(k)},
\end{equation}
where the weighting coefficient is $\omega_i = (1-p_i^o(\varnothing))^2$. 
Note that, with this formulation, we effectively reduce the contributions of outputs that bring small information about the old classes ($p_i^o(\varnothing) \approx 1$), while we increase the contribution of important ones (where $p_i^o(\varnothing) \approx 0$). 

\parag{Mask-based Pseudo-labeling Strategy}
Forgetting is effectively reduced by the adaptive distillation loss but we are not, however, taking advantage of the presence of the old classes that are unlabeled in the dataset. We propose using the old model predictions to recognize old class segments and generate pseudo-labels to make efficient use of this information and improve the knowledge of past classes.

A simple strategy consists in considering the N output pairs coming from the old model independently and using all the pairs for which the predicted class is not $\varnothing$ as pseudo-labels. This strategy, however, does not consider two aspects: there may be pairs where the class is different from $\varnothing$ but (i) the mask substantially overlaps with a ground-truth segment, or (ii) the mask is noisy and has low confidence.

To overcome these issues, we propose a mask-based pseudo-labeling strategy that jointly considers the mask and class prediction confidences to avoid noisy labels and overlaps with the existing annotations. 
We define the model confidence as the multiplication among the class and mask probabilities. Formally, we compute the confidence for the $i$-th output as $q_i = p^{max}_i \cdot m^o_i \in [0,1]^{H \times W}$, where $p^{max}_i = \max_{k=1}^{K^{t-1}} p^o_i(k)$ and $m^o_i$ is the binary mask predicted by the old model. Note that we do not consider the $\varnothing$ class in the $\max$ operation. 
We denote the pseudo-class $c^{ps}_i = \argmax_{k=1}^{K^{t-1}} p^o_i(k)$ and we generate the pseudo-mask $m_i^{ps}$ considering two criteria: (i) there should be no overlap between the pseudo-mask and the ground-truth segments and (ii) to be included in a pseudo-mask, the confidence on a pixel should be maximum over all outputs.
Formally, denoting the binarization of the predicted mask as $m_i^{bin} = m_i > 0.5$ and the union of the ground-truth segments $m^{gt}_{all} = \sum_{i=1}^{N^{ps}} m_i^{gt}$, we generate the pseudo-mask as
\begin{equation} \small
    m_i^{ps}[h, w] = \begin{cases}
        m_i^{bin}[h, w]  & \text{if}\; i= \argmax_{j=1}^{N} q_j[h,w] \\ 
        & \;\;\;\; \text{and}\; m^{gt}_{all}[h,w] = 0,\\
        0 & \text{otherwise}.\\
    \end{cases}
\end{equation}

Indeed, not all pseudo-mask should be included in the set of pseudo-labels. There may be pseudo-masks where no pixel is active, \ie it is all zeros, or where the pseudo-mask contains only a small fraction of the original mask. For these reasons, we construct the set of pseudo-labels $z^{ps} = \{(c^{ps}_i, m^{ps}_i)\}_{i=1}^{N^{ps}}$ such that we include in $z^{ps}$ only the pseudo-masks that retain at least half of the pixels \wrt the binary mask $m_i^{bin}$ and that have at least one active pixel.

We denote the final annotation set $\bar{z} = \{\bar{c}_i, \bar{m}_i\}_{i=1}^{\bar{N}}$ as the union of the ground-truth labels $z^{gt}$ and the pseudo-labels $z^{ps}$, \ie $\bar{z} = z^{gt} \cup z^{ps}$ and $\bar{N}=N^{gt}+N^{ps}$. 

\parag{Overall training loss}
To train CoMFormer, a one-to-one matching $\sigma$ between the model predictions $z$ and the annotation set $\bar{z}$ is required. Following standard practices \cite{carion2020end, cheng2021masked}, we employ the Hungarian matching algorithm \cite{kuhn1955hungarianmatching} and we search for the matching $\sigma$ that minimize the assignment cost computed as $-p_i(\bar{c}_j) \cdot Dice(m_i, \bar{m}_j)$, where $Dice$ indicates the Dice coefficient \cite{dice1945diceloss}.
We assume the size of the prediction set to be larger than the annotations set (\ie $N > \bar{N}$) and to obtain a one-to-one matching we pad the annotations with “no object" tokens $\varnothing$. 

Once the best matching $\sigma$ has been found, we define the overall training loss as:
\begin{equation}
    \mathcal{L} = \mathcal{L}_{Seg}(z, \bar{z}) + \lambda_D \mathcal{L}_{AD}(p, p^o),
\end{equation}
where $\lambda_D$ is a trade-off hyper-parameter and
\begin{equation}
\begin{aligned}
    \mathcal{L}_{Seg}(z, \bar{z}) = \sum_{j=1}^{\bar{N}} [ - \alpha (1-p_{\sigma(j)}(\bar{c}_j))^\gamma \log p_{\sigma(j)}(\bar{c}_j) + \\ + \lambda_{mask} \mathds{1}_{\bar{c}_j \neq \varnothing} \mathcal{L}_{mask}(m_{\sigma(j)}, \bar{m}_j) ].
\end{aligned}
\end{equation}
We note that the first term in $\mathcal{L}_{Seg}$ indicates the focal loss \cite{lin2017focalloss}, where $\alpha$ and $\gamma$ are hyper-parameters.
Finally, $\mathcal{L}_{mask}$ is the sum of dice and cross-entropy losses, and $\lambda_{mask}$ is a hyper-parameter. We note that $\mathcal{L}_{mask}$ is computed only for valid segments, \ie where $\bar{c}_j \neq \varnothing$.

\section{Experiments}
\label{sec:exp}

\begin{table*}[t]
\centering
\small
\begin{tabular}{l|cccc|cccc|cccc}
\hline
\multicolumn{1}{c}{} & \multicolumn{4}{c}{{\textbf{100-50} (2 tasks)}}  & \multicolumn{4}{c}{{\textbf{100-10} (6 tasks)}} & \multicolumn{4}{c}{{\textbf{100-5} (11 tasks)}} \\
\textbf{Method}       & \textit{1-100} & \textit{101-150} & \textit{all} & \textit{avg} & \textit{1-100} & \textit{101-150}  & \textit{all} & \textit{avg}  & \textit{1-100} & \textit{101-150} & \textit{all} & \textit{avg}  \\ \hline
FT           &  0.0 & 25.8  & 8.6 & 24.8	&  0.0 &   2.9  &  1.0 & 7.9 &  0.0	&  1.3 	& 0.4 &  4.6  \\ 
MiB          & 35.1 & 19.3  &29.8 & 35.4	& 27.1 &  10.0  & 21.4 & 29.1 & 24.0	&  6.5  &	18.1& 25.6  \\ 
PLOP         & 41.0 & 26.6 &  36.2 & 38.6	& 30.5 &  \textbf{17.5} & 26.1 & 32.9  & 28.1	& 15.7  &	24.0& 30.5  \\ 
\textbf{CoMFormer}    & \textbf{41.1} & \textbf{27.7} & \textbf{36.7} & \textbf{38.8} &\textbf{36.0} &  17.1  & \textbf{29.7}& \textbf{35.3} & \textbf{34.4}	& \textbf{15.9}  &	\textbf{28.2}& \textbf{34.0} \\ \hdashline
\textit{Joint}        & \textit{43.2} & \textit{32.1}  & \textit{39.5} & ---  & \textit{43.2} &  \textit{32.1}   & \textit{39.5} & ---  & \textit{43.2}	& \textit{32.1}   &	\textit{39.5} & --- \\ 
\hline
\end{tabular}
\caption{\textbf{Continual Panoptic Segmentation} results on ADE20K dataset in PQ considering multiple training protocols.}
\label{tab:ade_pan}

\end{table*}

\begin{table*}[t]
\centering
\small
\setlength{\tabcolsep}{4pt} %
\begin{tabular}{ll|cccc|cccc|cccc}
\hline
\multicolumn{2}{c}{} & \multicolumn{4}{c}{{\textbf{100-50} (2 tasks)}} & \multicolumn{4}{c}{{\textbf{100-10} (6 tasks)}} & \multicolumn{4}{c}{{\textbf{100-5} (11 tasks)}} \\
\textbf{Backbone}   & \textbf{Method}       & \textit{1-100} & \textit{101-150} & \textit{all} & \textit{avg}  & \textit{1-100} & \textit{101-150}  & \textit{all} & \textit{avg} & \textit{1-100} & \textit{101-150}  & \textit{all} & \textit{avg}  \\ \hline
\multirow{3}{*}{DeepLab-v3 \cite{chen2017deeplabv3}} 
& MiB \cite{cermelli2020modelingthebackground}          & 40.5 & 17.2  &	32.8 & 37.3 & 38.3 &	11.3  & 29.2 &	35.1 & 36.0 &  5.7  & 26.0&	32.7 \\
& PLOP \cite{douillard2020plop}                         & 41.9 & 14.9  &	32.9 & 37.4  &40.5 &	14.1  & 31.6 &	36.6 & 39.1 &  7.8  & 28.8 &	35.3  \\
& RCIL \cite{zhang2022rcil}                             & 42.3 & 18.8   &	34.5 & --- &39.3 &	\textbf{17.6}  & 32.1 &---	  & 38.5 & 11.5   & 29.6 &	---  \\	 \hline
\multirow{2}{*}{Per-Pixel} & MiB          & 40.3 & 24.0  & 34.8 & 37.5  & 35.1 & 14.0  &	28.1 & 34.0 & 33.3 & \textbf{15.2}  &  27.3 & 33.1 \\ 
& PLOP        & 40.2 & 20.2  & 33.5  & 36.9 &  32.6 &	13.7  & 26.3 & 32.4 & 33.3 & 9.4  & 25.4 & 32.8\\ 
\hline
\multirow{5}{*}{Mask-based} & FT        &  0.0 & \textbf{26.7} &  8.9 & 26.4 &  0.0 &  2.3  &	 0.8 &	 8.5 &  0.0 &  1.1 &  0.3 &  4.2 \\
& MiB       & 37.0 & 24.1  &  32.6 & 38.3  & 23.5 & 10.6  &	26.6 &	29.6 & 21.0 &  6.1  & 16.1 & 27.7 \\
& PLOP     & 44.2 & 26.2  & 38.2 & 41.1  & 34.8 & 15.9  &	28.5 &	35.2 & 33.6 & 14.1  & 27.1 & 33.6 \\
& \textbf{CoMFormer}    & \textbf{44.7} & 26.2  & \textbf{38.4} & \textbf{41.2} &  \textbf{40.6} & 15.6  &	\textbf{32.3} &	\textbf{37.4} & \textbf{39.5} & 13.6  & \textbf{30.9} & \textbf{36.5} \\  \hdashline
& \textit{Joint}        & \textit{46.9} & \textit{35.6}   & \textit{43.1} & --- & \textit{46.9} & \textit{35.6}   & \textit{43.1} & --- & \textit{46.9} & \textit{35.6}   &	\textit{43.1} &	--- \\ 
\hline
\end{tabular}
\caption{\textbf{Continual Semantic Segmentation} results on the ADE20K dataset in mIoU considering multiple training protocols. We report the numbers published by the original papers on DeepLab-v3 \cite{chen2017deeplabv3}. The remaining results are based on our re-implementation (see \cref{sec:baselines}).}
\label{tab:ade_sem}
\vspace{-1em}
\end{table*}

\subsection{Dataset and Settings}
We start from the widely adopted continual semantic segmentation benchmark defined in \cite{cermelli2020modelingthebackground} and we extend it to the newly proposed continual panoptic segmentation using a modified Continuum library \cite{douillardlesort2021continuum}. 
We compare our novel model on the large-scale ADE20K \cite{zhou2017adedataset} dataset since it supports both semantic and panoptic segmentation tasks. This challenging dataset contains 150 classes, divided into 100 “things" and 50 “stuff" categories. This dataset represents a wide variety of scenes, both interior and exterior, with images featuring an average of 9.9 classes, while other datasets like COCO \cite{lin2014mscocodataset} only have an average of 3.5 classes.

\parag{Continual Learning Protocols}
Previous continual semantic segmentation works \cite{douillard2020plop, cermelli2020modelingthebackground} describe a training protocol to assess the performance on multiple continual learning steps. In particular, \cite{cermelli2020modelingthebackground} proposed three protocols for ADE20K with different numbers of tasks: (i) the \textit{100-50} consists of two tasks, the first of 100 and the second of 50 classes; (ii) the \textit{50-50} consists of three tasks of 50 classes; (iii) the \textit{100-10} consists on 6 tasks, the first on 100 classes followed by 5 tasks of 10 classes. In addition, we use the \textit{100-5} introduced in \cite{douillard2020plop}, consisting of 11 tasks, the first of 100 classes followed by 10 tasks of 5 classes.
To divide the images into different tasks, we follow the splits provided by \cite{cermelli2020modelingthebackground} for \textit{100-50}, \textit{100-10}, \textit{50-50} for both semantic and panoptic segmentation. In particular, they ensured that each image appears only on a unique task. Differently, for the \textit{100-5} we use the split proposed by \cite{douillard2020plop}, where the same image may appear on multiple tasks. 
We report the results of the \textit{50-50} setting for both semantic and panoptic segmentation in the supplementary material.

\parag{Metrics}
We compare the different models using the mean Intersection over Union (mIoU) on semantic segmentation \cite{everingham2015pascalvoc} and Panoptic Quality (PQ) on panoptic segmentation \cite{kirillov2019panoptic}. PQ is the product of two components: Segmentation Quality (SQ) which considers the IoU between the correctly classified segments, and Recognition Quality (RQ) which only considers the classification accuracy. For both tasks, we report the metric after the last step $T$ on the first classes ($C^1$), for the added classes ($C^{2:T}$), and for all the classes (\textit{all}). We also report the average of the final performance on all seen classes after each step (\textit{avg}) following \cite{douillard2020plop}.

\subsection{Baselines} \label{sec:baselines}
We benchmark our model against the state-of-the-art methods: MiB \cite{cermelli2020modelingthebackground}, PLOP \cite{douillard2020plop}, and RCIL \cite{zhang2022rcil}. In semantic segmentation, we report them with the original DeepLab-v3 segmentation architecture \cite{chen2017deeplabv3}. In addition, for a fair comparison, we report MiB and PLOP using the CoMFormer architecture. In particular, we implement the methods in two versions: (i) we replace the CoMFormer losses with a per-pixel cross-entropy loss on the dot product between mask and class logits and we apply the continual learning methods on that baseline (we refer to this as \textit{Per-Pixel}), and (ii) by applying the distillation losses directly on the classification head of the CoMFormer architecture (\textit{Mask-based}). We note that the first version cannot be applied to panoptic segmentation, since the task does not support the use of per-pixel losses. For PLOP, we apply its distillation loss local-POD on the intermediate features of the segmentation backbone, as in the original paper, and we use the pseudo-labeling strategy proposed in \cref{sec:method}. RCIL was not re-implemented since it is not possible to extend their approach to non-convolutional architectures like ours. We include a naive finetuning (FT) without any capabilities for continual learning. 
Finally, we also report an upper bound (\textit{Joint}) that is trained in the traditional segmentation setting. We made our best effort to find the most suitable hyper-parameters for all the baselines. 

\subsection{Implementation Details}
\vspace{-1em}\parag{Architecture} Following the previous benchmarks \cite{cermelli2020modelingthebackground, douillard2020plop} we use as backbone a ResNet101 \cite{he2016resnet} on semantic segmentation and we propose to use a ResNet50 \cite{he2016resnet} for panoptic segmentation. For both tasks, we use the pixel decoder and the transformer decoder proposed in \cite{cheng2021masked} with 9 layers in total and $N=100$ queries. In addition, we adopt all the improvements introduced in \cite{cheng2021masked}: masked attention, multi-scale features, and their optimization improvements. 
We report the performance of single-scale inference.

\parag{Training parameters}
We follow the Mask2Former \cite{cheng2021masked} hyper-parameters and we use AdamW \cite{ilya2019decoupled} optimizer with an initial learning rate of $0.0001$ for the first step ($t=0$) and $0.00005$ in the following ($t>0$). A learning rate multiplier of 0.1 is applied to the backbone and we follow a polynomial learning rate schedule. We use weight decay of $0.05$. We train the model for 160K iterations in the first step and 400 iterations per class in the following (e.g. learning 50 classes we train for $50 \times 400 = 20000$ iterations).
For semantic segmentation, we use a crop size of 512 and a batch size of 16. For panoptic segmentation, we use a crop size of 640 and a batch size of 8. For both, we use the standard random scale jittering between 0.5 and 2.0, random horizontal flipping, random cropping, as well as random color jittering as data augmentation \cite{cheng2021maskformer}. We set $\alpha=20$, $\gamma=2$, and $\lambda_{mask}=5.0$. $\lambda_D$ is set to $1$ for \textit{100-50} and to $10$ in the \textit{100-10} and \textit{100-5}.
Finally, following the standard protocol of continual learning \cite{cermelli2020modelingthebackground, douillard2020plop} we do not store any image of previous steps and we do not use rehearsal learning.

\begin{figure*}
    \includegraphics[width=0.97\textwidth]{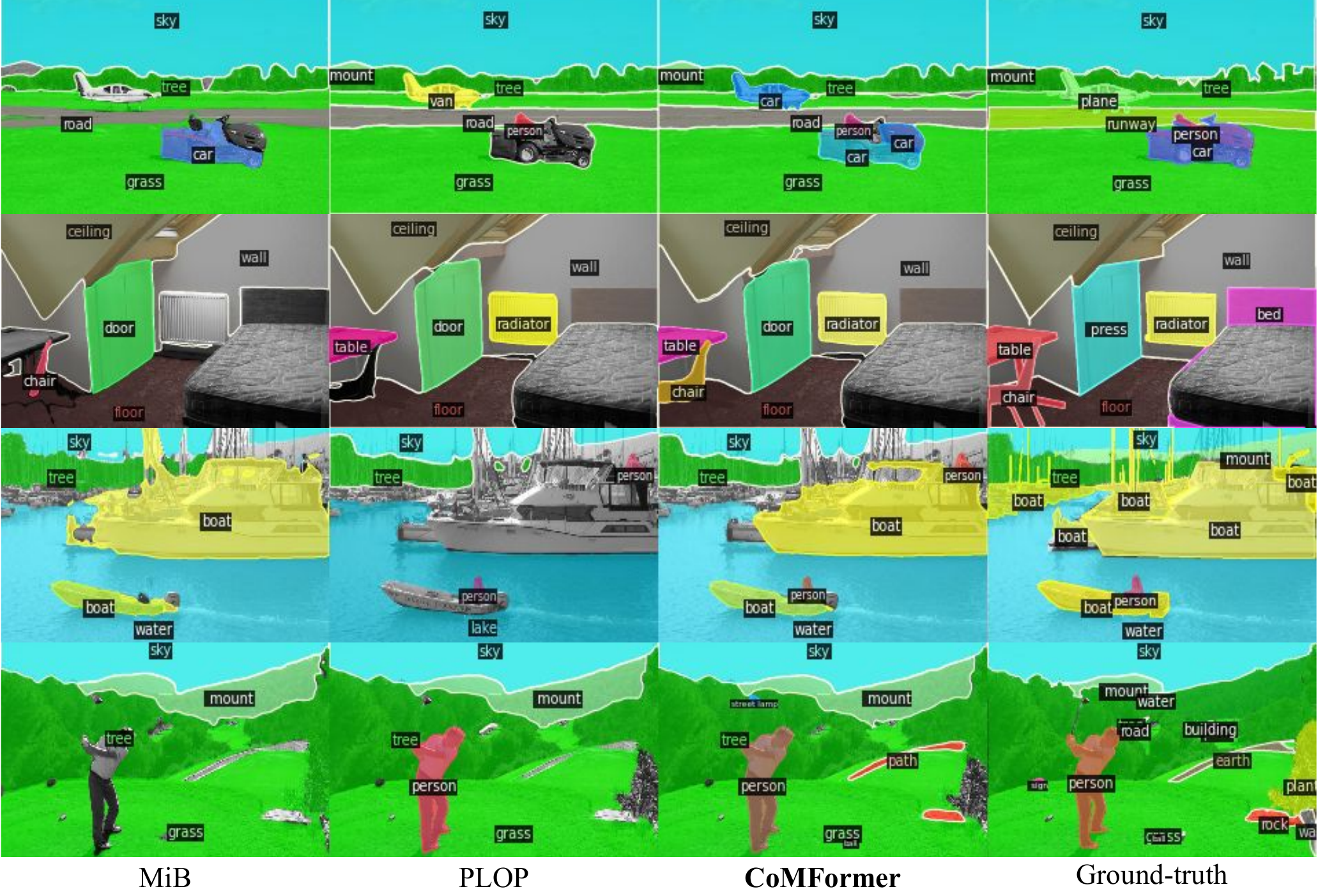}
    \caption{\textbf{Qualitative results} of CoMFormer \textit{v.s.} MiB and PLOP on the \textit{100-10} continual panoptic segmentation setting on ADE20K.}
    \label{fig:qualitative_results} \vspace{-1em}
\end{figure*}

\subsection{Continual Panoptic Segmentation}
\cref{tab:ade_pan} reports the results on the new continual panoptic segmentation benchmark. CoMFormer exceeds all the baselines by a significant margin. %
Considering the \textit{100-50}, we can see that CoMFormer outperforms MiB on both old (+6 PQ) and new (+8.4 PQ) classes and PLOP on new classes (+1.1 PQ). Considering a longer sequence of tasks (\textit{100-10} and \textit{100-5}), we can see that the gap between CoMFormer and the other methods becomes more significant, especially in the old classes. In \textit{100-10}, it surpasses MiB and PLOP respectively by 8.9 and 5.5 PQ in old classes, while in the new classes it exceeds MiB by 7.1 and is comparable to PLOP. In the \textit{100-5}, it outperforms MiB and PLOP on both old (+10.4 and +6.3 PQ) and new classes (+9.4 and +0.2 PQ). Finally, we note that CoMFormer obtains performance close to the \textit{Joint} baseline in the \textit{100-50} setting, suffering a drop of only 2.1 PQ on the old classes and 4.4 PQ on the new ones. In the other settings, the gap is still relevant, indicating that learning from multiple training steps is still a very challenging task. 

In \cref{fig:qualitative_results}, we report a qualitative comparison with the baselines in the \textit{100-10} setting using randomly chosen images. We can see that our method is able to maintain better the previous knowledge by precisely segmenting all the classes in the image; for example, \wrt to PLOP, our method correctly segments the \textit{chair} in the second row and the \textit{boat} in the third row. We can note a common error: models correctly segment but misclassify objects in the image, like the \textit{plane} classified as \textit{car} in the first row or the \textit{press} classified as a \textit{door} in the third row. 

\subsection{Continual Semantic Segmentation} 
We report the comparison with state-of-the-art methods in continual semantic segmentation in \cref{tab:ade_sem}. 
CoMFormer achieves the best results on every setting, surpassing both methods based on DeepLab-v3 \cite{chen2017deeplabv3} and baselines implemented upon the same architecture. In particular, we can see that CoMFormer outperforms the best state-of-the-art method (RCIL \cite{zhang2022rcil} by 3.9\% on the \textit{100-50}, by 0.2\% on \textit{100-10}, and by 1.3\% on the \textit{100-5}. Moreover, note that we always improve the performance on the old classes and the \textit{avg} across different learning steps. 
In addition, CoMFormer exceeds all the implemented baselines. Comparing CoMFormer with PLOP (Mask-based), we can see that it obtains a substantial performance improvement, especially considering scenarios with longer training protocols: CoMFormers achieves 30.9\% (32.3\%) on the \textit{all} metrics, while PLOP only 27.1\% (28.5\%) on the \textit{100-5} (\textit{100-10}) setting.
Furthermore, the results confirm that improvements are not led by the newer architecture. Per-Pixel baselines achieve only similar or even lower performance than their counterparts. MiB (Per-Pixel), for example, slightly improves the one based on DeepLab-v3 on the \textit{100-50} (+2.0\% on \textit{all}) and in the \textit{100-5} (+1.3\%) but its performance is slightly worse in the \textit{100-10} (-1.1\%). In addition, note that Per-Pixel baselines demonstrate a large performance drop in the old classes \wrt their counterpart, showing that, without employing a distillation strategy tailored to the architecture, it is difficult to maintain satisfactory results.

\subsection{Ablation Studies}
\begin{table}[t]
    \centering
    \small
    \setlength{\tabcolsep}{3pt} %
    \begin{tabular}{l|ccc|ccc}
    \hline
\multicolumn{1}{c}{} & \multicolumn{3}{c}{\textbf{Semantic} - mIoU} & \multicolumn{3}{c}{\textbf{Panoptic} - PQ}\\
\textbf{Method} & \textit{1-100} & \textit{101-150} & \textit{all} & \textit{1-100} & \textit{101-150} & \textit{all}\\ \hline
\multicolumn{1}{c|}{} & \multicolumn{6}{c}{\textbf{100-50} (2 tasks)} \\  \hline
Mask2Former & 44.3	& \textbf{26.0} & 38.1	&       37.9 &	24.5	&	33.4 \\
CoMFormer   & \textbf{44.7}	& \textbf{26.2} & \textbf{38.4}	&       \textbf{41.1} &	\textbf{27.7}	&	\textbf{36.7} \\ \hline
\multicolumn{1}{c|}{} & \multicolumn{6}{c}{\textbf{100-10} (6 tasks)} \\  \hline
Mask2Former & 38.7 &	\textbf{15.7}	& 31.0 &	30.2	& 10.8		& 23.7 \\
CoMFormer   & \textbf{40.6} &	\textbf{15.6}	& \textbf{32.3} &	\textbf{36.0}	& \textbf{17.1}		& \textbf{29.7} \\ \hline
\multicolumn{1}{c|}{} & \multicolumn{6}{c}{\textbf{100-5} (11 tasks)}  \\  \hline
Mask2Former & 32.6	& \textbf{13.7} &	26.3 &	25.9	&  5.6 &	19.1 \\
CoMFormer   &\textbf{ 39.5}	& \textbf{13.6} &	\textbf{30.9} &	\textbf{34.4}	& \textbf{15.9} &	\textbf{28.2} \\ \hline
\end{tabular}
\caption{\textbf{Comparison between CoMFormer and Mask2Former} \cite{cheng2021masked} in continual semantic and panoptic segmentation. We apply to Mask2Former our mask-based pseudo-labeling strategy and the adaptive distillation loss for a fair comparison.} %
\label{tab:ablation_m2f}
\end{table}

\vspace{-1em} \parag{Mask2Former vs CoMFormer}
In \cref{tab:ablation_m2f}, we compare the proposed CoMFormer against Mask2Former \cite{cheng2021masked}. In particular, to fairly compare the two methods, we add our adaptive distillation loss and the pseudo-labeling strategy to Mask2Former. Considering semantic segmentation, CoMFormer achieves better results on the old classes (on average 3\% mIoU), while the two methods obtain comparable results on the new classes. On panoptic segmentation, CoMFormer exceeds Mask2Former in all the classes with a considerable gap (6 PQ on both old and new classes on average). We conclude that a vanilla MaskFormer architecture is not suited to continual learning because it suffers from forgetting, while CoMFormer outperforms it in any setting.

\begin{table}
\centering
    \small
    \setlength{\tabcolsep}{3pt} %
\begin{tabular}{ccc|ccc|ccc}
    \hline
\multicolumn{3}{c}{} & \multicolumn{3}{c}{\textbf{100-50} (2 tasks) } & \multicolumn{3}{c}{\textbf{100-10} (6 tasks)} \\
PS & KD & AD & \textit{1-100} & \textit{101-150} & \textit{all} & \textit{1-100} & \textit{101-150} & \textit{all}\\
\hline
- & - & - &  0.0 & 25.8 &  8.6 &  0.0 &  2.9 &  1.0 \\
- & - & \checkmark &  28.4 &	15.3 &	24.0 & 23.0 &  9.1 & 18.4 \\ 
\checkmark & - & - & 40.7 & 26.3 & 35.9 & 30.1 & 17.2 & 25.8 \\
\checkmark & \checkmark & - & 40.5 & \textbf{28.1} & 36.4 & 30.2 & \textbf{17.6} & 26.0 \\
\checkmark & - & \checkmark & \textbf{41.1} & 27.7 & \textbf{36.7} & \textbf{36.0} & 17.1 & \textbf{29.7} \\
\hline
\end{tabular}
\caption{\textbf{Ablation of the different key components of our CoMFormer architecture}. We report the PQ on the 100-50 and 100-10 panoptic settings. PS: Pseudo-labeling; AD: adaptive distillation \cref{eq:qd}; KD: standard distillation \cref{eq:ukd}} \vspace{-1em}
\label{tab:ablation}
\end{table}

\parag{Method components}
We investigate the benefit of each method component in \cref{tab:ablation}, reporting the results in PQ on the \textit{100-50} and \textit{100-10} settings in panoptic segmentation. Without applying any regularization technique, the performance on old classes drops to 0, denoting that it is important to deal with forgetting. 
Both the adaptive distillation loss (AD) and the pseudo-labeling strategy (PS) lead to improvement in the old classes when applied independently. However, we obtain the best results when combining them, especially considering the \textit{100-10} scenario, where the old classes are improved by 5.8 PQ and the final performances of 3.9 PQ. Finally, the table shows that weighting the queries based on their semantic content is essential to improve performance: comparing a standard distillation (KD \cref{eq:ukd}) with our adaptive distillation (AD \cref{eq:qd}), we largely improve the performance on both settings: +0.3 PQ in \textit{100-50} and +3.7 PQ in \textit{100-10} considering all classes.

\begin{table}[t]
    \centering
    \small
    \begin{tabular}{l|ccc|ccc}
    \hline
\multicolumn{1}{c}{} & \multicolumn{3}{c}{\textbf{100-50} (2 tasks) } & \multicolumn{3}{c}{\textbf{100-10} (6 tasks)} \\
\textbf{Method}       & \textit{RQ} & \textit{SQ} & \textit{PQ} & \textit{RQ} & \textit{SQ} & \textit{PQ} \\ \hline
MiB       &  36.8 & 73.0 & 29.8 & 26.4 & 55.5 & 21.4 \\
PLOP      &  43.8 & 79.1 & 36.2 & 31.7 & 61.1 & 26.1 \\
\textbf{CoMFormer} & \textbf{44.4} & \textbf{79.6} & \textbf{36.7} & \textbf{36.2} & \textbf{73.4} & \textbf{29.7} \\ \hdashline
\textit{Joint}     &  47.6 & 79.0 & 39.5 & 47.6 & 79.0 & 39.5 \\
\hline
    \end{tabular}
    \caption{\textbf{Panoptic metrics analysis}. We report Recognition Quality (RQ), Segmentation Quality (SQ), and Panoptic Quality (PQ) over all the classes after the last step.} \vspace{-1.2em}
    \label{tab:ablation_forg}
\end{table}

\parag{Forgetting in CoMFormer} \label{sec:ablation}
To investigate the reason for the forgetting, we considered separately the Recognition Quality (RQ) and the Segmentation Quality (SQ) in panoptic segmentation. RQ measures how well the model is able to classify segments, while SQ considers the precision of the segmentation. Note that, as reported in \cite{kirillov2019panoptic}, SQ and RQ are not independent since SQ is computed only on segments correctly classified. 
The results on the \textit{100-50} and \textit{100-10} are reported in \cref{tab:ablation_forg}. The table clearly shows that, as hypothesized in \cref{sec:lwf}, forgetting happens heavily on the classification ability while it is less marked on the segmentation ability of the model. CoMFormer, in fact, without providing regularization losses on the mask prediction, is able to match the segmentation ability of the \textit{Joint} upper bound on the \textit{100-50} and it gets close results on the \textit{100-10} (-5.6). In addition, it is evident that classification ability is heavily affected by forgetting. CoMFormer has a gap with \textit{Joint} of 3.2 and 11.4 RQ respectively in the \textit{100-50} and \textit{100-10}. We note that, however, other baselines are more affected by forgetting, showcasing even larger gaps.

\vspace{-0.3em} 
\section{Conclusion} %
\label{sec:conclusion}
\vspace{-0.5em}
In this paper, we extend the task of continual learning beyond semantic segmentation by introducing the challenging and realistic continual panoptic segmentation setting. We propose CoMFormer, a strategy to address segmentation as a mask classification problem exploiting the properties of transformers to tackle both semantic and panoptic segmentation tasks. 
The extensive quantitative results highlight that CoMFormer outperforms all previous methods, both in semantic and panoptic segmentation, under a unified paradigm. In future work, we aim to extend the benchmark to instance segmentation to show that CoMFormer can be employed to address any continual segmentation task. 

{ \small \textbf{Acknowledgments.}
Fabio Cermelli acknowledges travel support from ELISE (GA no 951847). We acknowledge that the research activity  was carried out using the IIT HPC infrastructure.}

{\small
\bibliographystyle{ieee_fullname}
\bibliography{egbib,arthurcitations}
}

\clearpage
\section*{Appendix}
\appendix
\section{Additional quantitative results}

\parag{50-50 in Continual Panoptic Segmentation}
In \cref{tab:ade_pan_50} we report additional experiments on Continual Panoptic Segmentation on the \textit{50-50} setting where we perform three tasks of 50 classes.
CoMFormer outperforms all the baselines, obtaining the best results on both old and new classes. In particular, we can see that it exceeds the best competitor, PLOP, by 0.5 PQ in the old classes and 0.2 PQ in the new ones. When comparing with MiB, however, we can see that the gap is more relevant: +11.6 PQ on old classes and +10.2 on the new ones. Finally, we can see that CoMFormer obtains a small performance gap with the \textit{Joint} baselines, which is more relevant for the new classes (-7.6 PQ).

\begin{table}[t]
    \centering
    \begin{tabular}{l|cccc}
    \hline
\multicolumn{1}{c}{} & \multicolumn{4}{c}{{\textbf{50-50} (11 tasks)}} \\
\textbf{Method}       & \textit{1-50} & \textit{51-150} & \textit{avg} & \textit{all}  \\ \hline
FT                    &  0.0 &	14.3 &	23.1 & 	9.5  \\
MiB                   & 33.6 &	16.3 &	31.8 & 	22.1  \\
PLOP                  & 44.7 &	26.3 &	\textbf{37.9} & 	32.4  \\
\textbf{CoMFormer}    &\textbf{45.2} &	\textbf{26.5} &	\textbf{37.9} & 	\textbf{32.7}  \\ \hdashline
\textit{Joint}        & 50.2 &	34.1 &	---	 &  39.5 \\
\hline
    \end{tabular}
    \caption{Continual Panoptic Segmentation results on ADE20K dataset on 50-50 setting in PQ.}
    \label{tab:ade_pan_50}
\end{table}

\begin{table}[t]
    \centering
    \small
    \setlength{\tabcolsep}{4pt} %
    \begin{tabular}{ll|cccc}
    \hline
\multicolumn{2}{c}{} & \multicolumn{4}{c}{{\textbf{50-50} (11 tasks)}} \\
\textbf{Architecture}       & \textbf{Method}       & \textit{1-50} & \textit{51-150} & \textit{avg} & \textit{all}  \\ \hline
\multirow{3}{*}{DeepLab-v3 \cite{chen2017deeplabv3}} 
& MiB \cite{cermelli2020modelingthebackground}  & 45.3 & 21.6 &	38.9 & 29.3 \\
& PLOP \cite{douillard2020plop}                 & 48.6 & 21.6 &	39.4 & 30.4 \\    
& RCIL \cite{zhang2022rcil}                     & 48.3 & 25.0 &	---  & 32.5 \\   \hline 
\multirow{2}{*}{Per-Pixel}
& MiB                     & 44.9 & 25.4 &	35.0 & 31.9  \\
& PLOP                    & 43.2 & 24.7 &	34.6 & 30.9  \\ \hline 
\multirow{5}{*}{Mask-Based}
& FT                      &  0.0	& 13.3 & 12.8 &	 8.9  \\
& MiB                     & 24.6	& 19.4 & 25.8 &	21.1  \\
& PLOP                    & 48.1	& \textbf{26.6} & 36.5 &	33.8  \\
& \textbf{CoMFormer}      & \textbf{49.2}	& \textbf{26.6} & \textbf{36.6} &	\textbf{34.1}  \\ \hdashline
& \textit{Joint}          & 53.4	& 38.0 &  --- &	43.1  \\
\hline
    \end{tabular}
    \caption{Continual Semantic Segmentation results on ADE20K dataset on 50-50 setting in mIoU.}
    \label{tab:ade_sem_50}
\end{table}

\parag{50-50 in Continual Semantic Segmentation}
\cref{tab:ade_sem_50} reports the additional results on the Continual Semantic Segmentation benchmark on the \textit{50-50} setting in mIoU, comparing CoMFormer with previous works based on DeepLab \cite{chen2017deeplabv3} and our re-implementation based on the CoMFormer architecture, both in Per-Pixel and Mask-Based fashion.
We observe that CoMFormer achieves a new state of the art. In particular, when comparing it with previous works, we can see that it outperforms the best baseline (RCIL) on both old (+0.9 mIoU) and new classes (+1.6 mIoU), for an overall improvement of 1.6 mIoU. Furthermore, CoMFormer also outperforms the baselines implemented on the same architecture: \wrt to Per-Pixel baselines, there is a relevant performance gap, especially regarding the old classes (CoMFormer 49.2 vs MiB 44.9 mIoU). Considering the Mask-Based baselines, CoMFormer shows the best performance, improving PLOP by 1.1 mIoU on the old classes and by 0.3 mIoU on all.

\begin{figure*}
    \centering
    \includegraphics[width=0.9\textwidth]{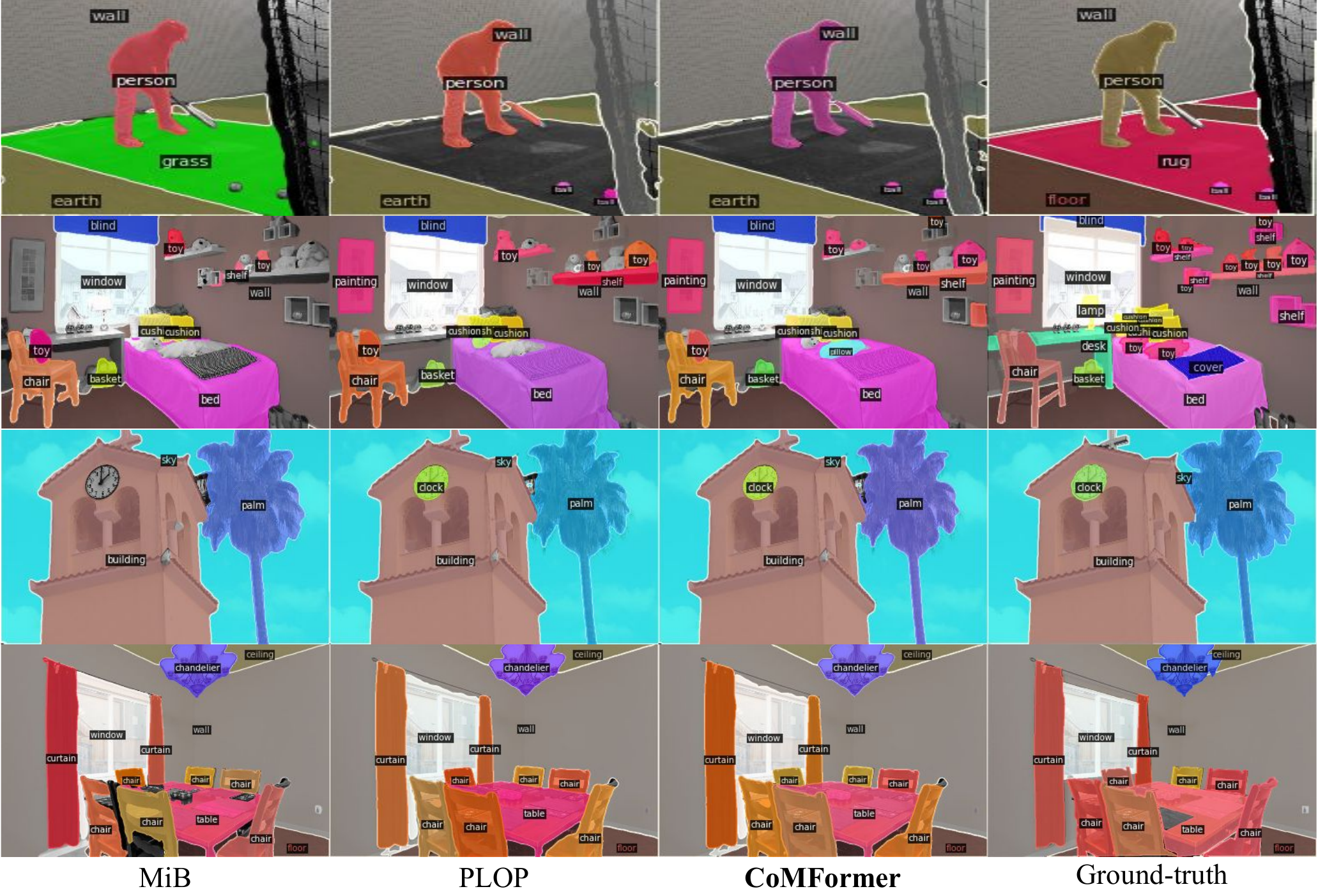}
    \caption{\textbf{Qualitative results} of CoMFormer \textit{v.s.} MiB and PLOP on the \textit{100-50} continual panoptic segmentation setting on ADE20K.}
    \label{fig:qr_P10050} \vspace{-1em}
\end{figure*}
\begin{figure*}
    \centering
    \includegraphics[width=0.9\textwidth]{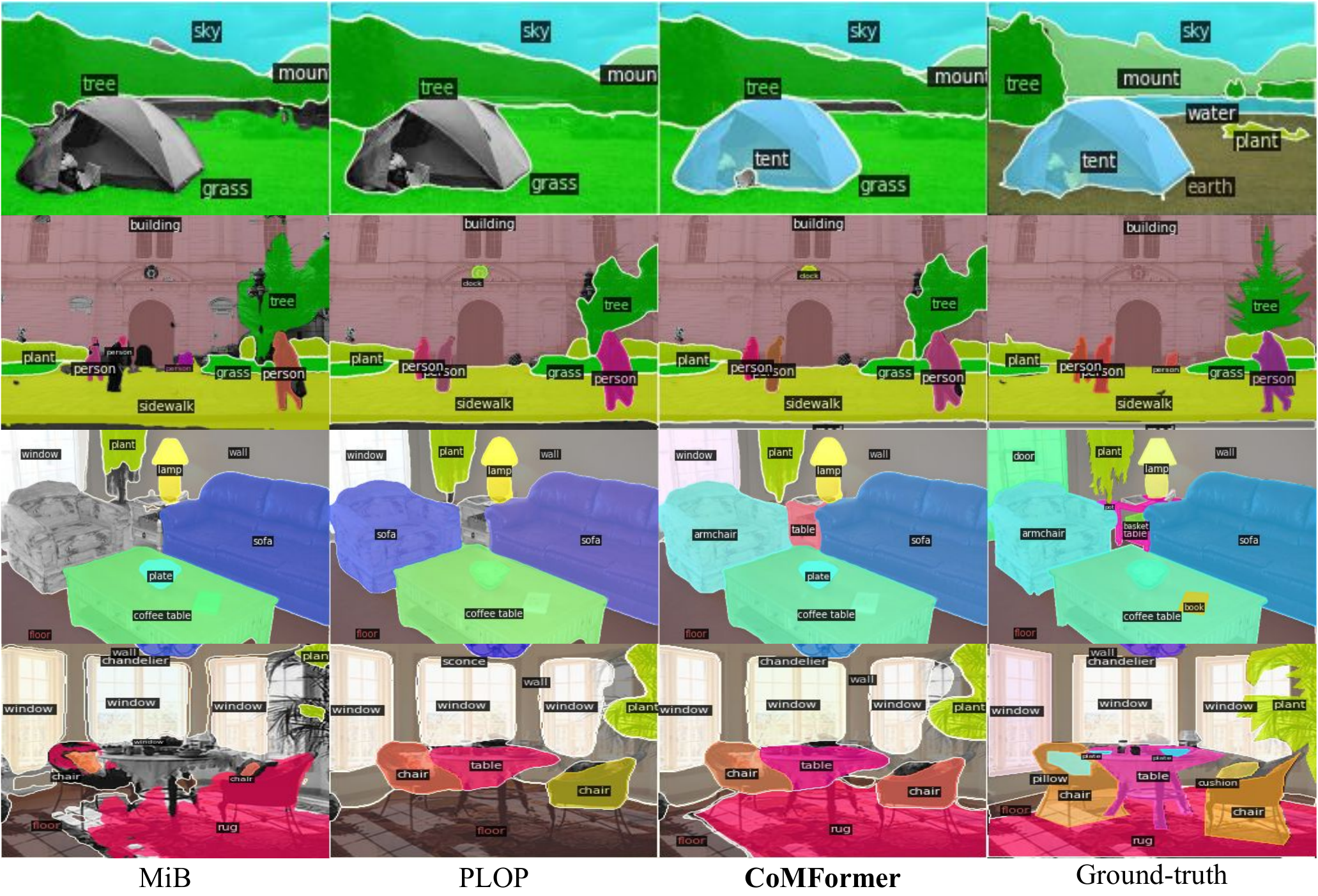}
    \caption{\textbf{Qualitative results} of CoMFormer \textit{v.s.} MiB and PLOP on the \textit{100-5} continual panoptic segmentation setting on ADE20K.}
    \label{fig:qr_P1005}  \vspace{-1em}
\end{figure*}
\begin{figure*}
    \centering
    \includegraphics[width=\textwidth]{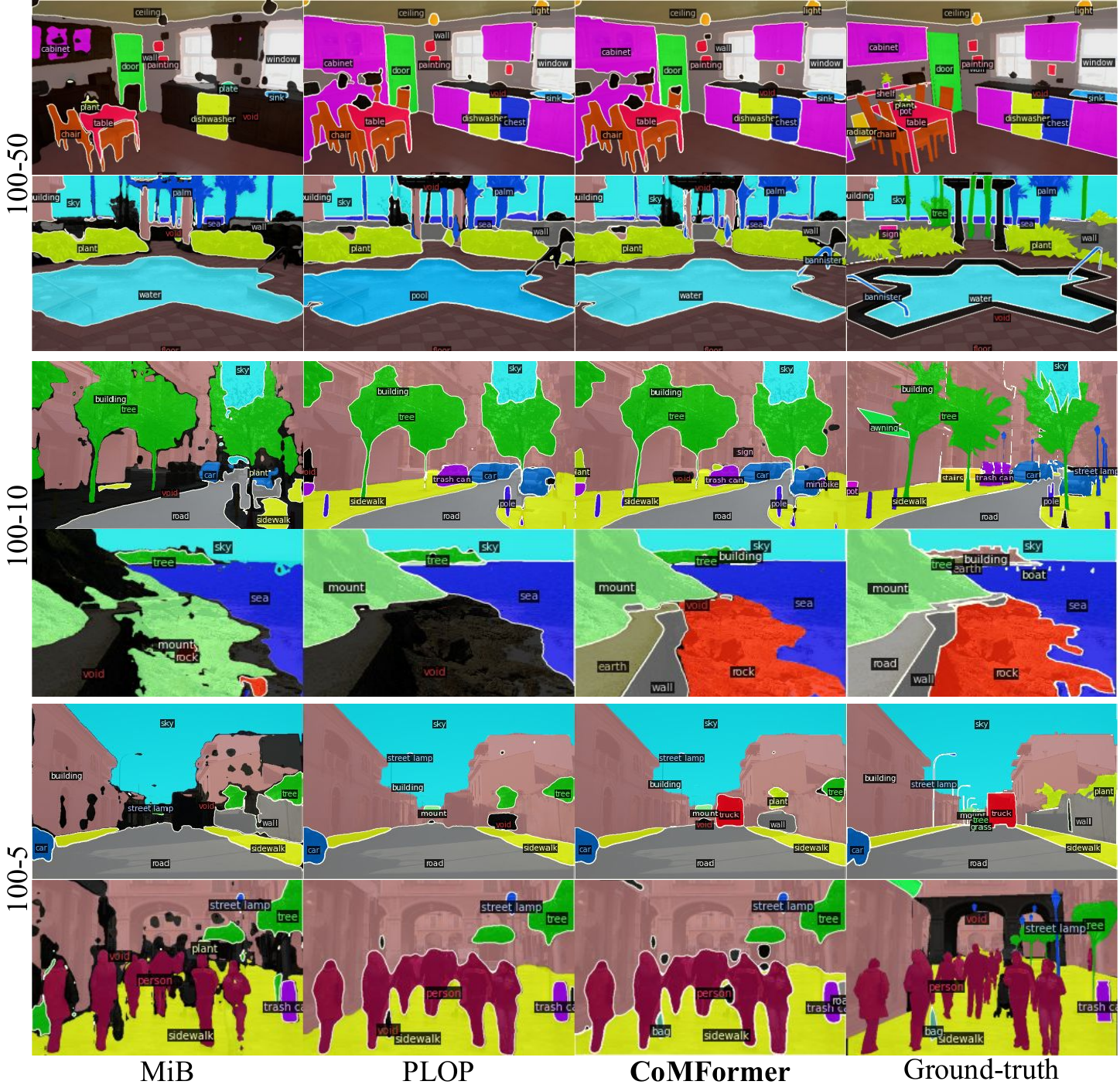}
    \caption{\textbf{Qualitative results} of CoMFormer \textit{v.s.} MiB and PLOP on multiple settings of the continual semantic segmentation benchmark on ADE20K.}
    \label{fig:qualitative_results} \vspace{-1em}
    \label{fig:qr_Seg} 
\end{figure*}

\section{Additional qualitative results}
\parag{Continual Panoptic Segmentation}
\cref{fig:qr_P10050} and \cref{fig:qr_P1005} report additional qualitative results on, respectively, the \textit{100-50} and \textit{100-5} settings in continual panoptic segmentation, comparing CoMFormer with MiB and PLOP using images randomly sampled from the validation set. Considering the \textit{100-50}, we can see that PLOP and CoMFormer achieve visually similar results, while MiB struggles in segmenting every image object (for example, the \textit{clock} in the third row). 
Differently, on the \textit{100-5}, CoMFormer visually outperforms the other baselines being able to correctly segment all the objects in the image (\eg the \textit{tent} in the first row, the \textit{table} in the third row, and the \textit{rug} in the fourth row). However, we note a common error across all the methods: some classes are correctly segmented but misclassified (\eg \textit{grass} instead of \textit{earth} in the first row and \textit{window} instead of \textit{door} in the third). This error is less present in CoMFormer \wrt PLOP and MiB, as can be seen from the \textit{chandelier} in the last row and the \textit{armchair} in the third.

\parag{Continual Semantic Segmentation}
\cref{fig:qr_Seg} reports the qualitative results for the \textit{100-50}, \textit{100-10}, and \textit{100-5} settings of the continual semantic segmentation benchmark comparing CoMFormer with MiB and PLOP on images randomly sampled from the validation set. 
Considering the \textit{100-50} setting, MiB is far worse than other baselines: it is not able to correctly segment the object in the image, achieving low performance. PLOP and CoMFormer achieve similar results, being able to segment all the objects in the images. 
Differently, on the \textit{100-10} setting, the difference among methods becomes more evident: considering the second row, CoMFormer correctly segments the \textit{rock} and the \textit{wall}, while misclassifying the \textit{road} with \textit{earth}. However, both PLOP and MiB are not able to segment the image: the former is not able to report any segment in that area, while the latter segments incorrectly the area as \textit{mount}.
Finally, considering the \textit{100-5} setting, we note that MiB achieves poor performance on both images, being unable to finely segment the image pixels. Comparing CoMFormer with PLOP, our model CoMFormer is able to segment more classes (\eg the \textit{truck} in the fifth row and the \textit{bag} in the last row), obtaining better performances.
Overall, the qualitative results confirm the quantitative findings, where CoMFormer outperforms the other methods, especially considering settings where multiple learning steps are performed. Those longer continual settings are more realistic and allows us to benchmark more efficiently what a truly lifelong learning agent should be.

\begin{table*}[t]
\centering
\begin{tabular}{rlc|rlc|rlc}
\textbf{Idx} & \multicolumn{1}{c}{\textbf{Name}} & \multicolumn{1}{c}{\textbf{Thing}} & \textbf{Idx} & \multicolumn{1}{c}{\textbf{Name}} & \textbf{Thing} & \textbf{Idx} & \multicolumn{1}{c}{\textbf{Name}} & \multicolumn{1}{c}{\textbf{Thing}} \\ \hline
1            & wall                              &                                    & 51           & refrigerator                      & \checkmark     & 101          & poster                            & \checkmark                         \\
2            & building                          &                                    & 52           & grandstand                        &                & 102          & stage                             &                                    \\
3            & sky                               &                                    & 53           & path                              &                & 103          & van                               & \checkmark                         \\
4            & floor                             &                                    & 54           & stairs                            & \checkmark     & 104          & ship                              & \checkmark                         \\
5            & tree                              & \checkmark                         & 55           & runway                            &                & 105          & fountain                          & \checkmark                         \\
6            & ceiling                           &                                    & 56           & case                              & \checkmark     & 106          & conveyer                          &                                    \\
7            & road                              &                                    & 57           & pool                              & \checkmark     & 107          & canopy                            & \checkmark                         \\
8            & bed                               & \checkmark                         & 58           & pillow                            & \checkmark     & 108          & washer                            & \checkmark                         \\
9            & windowpane                        & \checkmark                         & 59           & screen                            & \checkmark     & 109          & plaything                         & \checkmark                         \\
10           & grass                             &                                    & 60           & stairway                          &                & 110          & swimming                          &                                    \\
11           & cabinet                           & \checkmark                         & 61           & river                             &                & 111          & stool                             & \checkmark                         \\
12           & sidewalk                          &                                    & 62           & bridge                            &                & 112          & barrel                            & \checkmark                         \\
13           & person                            & \checkmark                         & 63           & bookcase                          & \checkmark     & 113          & basket                            & \checkmark                         \\
14           & earth                             &                                    & 64           & blind                             & \checkmark     & 114          & waterfall                         &                                    \\
15           & door                              & \checkmark                         & 65           & coffee                            & \checkmark     & 115          & tent                              & \checkmark                         \\
16           & table                             & \checkmark                         & 66           & toilet                            & \checkmark     & 116          & bag                               & \checkmark                         \\
17           & mountain                          &                                    & 67           & flower                            & \checkmark     & 117          & minibike                          & \checkmark                         \\
18           & plant                             & \checkmark                         & 68           & book                              & \checkmark     & 118          & cradle                            & \checkmark                         \\
19           & curtain                           & \checkmark                         & 69           & hill                              &                & 119          & oven                              & \checkmark                         \\
20           & chair                             & \checkmark                         & 70           & bench                             & \checkmark     & 120          & ball                              & \checkmark                         \\
21           & car                               & \checkmark                         & 71           & countertop                        & \checkmark     & 121          & food                              & \checkmark                         \\
22           & water                             &                                    & 72           & stove                             & \checkmark     & 122          & step                              & \checkmark                         \\
23           & painting                          & \checkmark                         & 73           & palm                              & \checkmark     & 123          & tank                              & \checkmark                         \\
24           & sofa                              & \checkmark                         & 74           & kitchen                           & \checkmark     & 124          & trade                             & \checkmark                         \\
25           & shelf                             & \checkmark                         & 75           & computer                          & \checkmark     & 125          & microwave                         & \checkmark                         \\
26           & house                             &                                    & 76           & swivel                            & \checkmark     & 126          & pot                               & \checkmark                         \\
27           & sea                               &                                    & 77           & boat                              & \checkmark     & 127          & animal                            & \checkmark                         \\
28           & mirror                            & \checkmark                         & 78           & bar                               & \checkmark     & 128          & bicycle                           & \checkmark                         \\
29           & rug                               &                                    & 79           & arcade                            & \checkmark     & 129          & lake                              &                                    \\
30           & field                             &                                    & 80           & hovel                             &                & 130          & dishwasher                        & \checkmark                         \\
31           & armchair                          & \checkmark                         & 81           & bus                               & \checkmark     & 131          & screen                            & \checkmark                         \\
32           & seat                              & \checkmark                         & 82           & towel                             & \checkmark     & 132          & blanket                           & \checkmark                         \\
33           & fence                             & \checkmark                         & 83           & light                             & \checkmark     & 133          & sculpture                         & \checkmark                         \\
34           & desk                              & \checkmark                         & 84           & truck                             & \checkmark     & 134          & hood                              & \checkmark                         \\
35           & rock                              & \checkmark                         & 85           & tower                             &                & 135          & sconce                            & \checkmark                         \\
36           & wardrobe                          & \checkmark                         & 86           & chandelier                        & \checkmark     & 136          & vase                              & \checkmark                         \\
37           & lamp                              & \checkmark                         & 87           & awning                            & \checkmark     & 137          & traffic                           & \checkmark                         \\
38           & bathtub                           & \checkmark                         & 88           & streetlight                       & \checkmark     & 138          & tray                              & \checkmark                         \\
39           & railing                           & \checkmark                         & 89           & booth                             & \checkmark     & 139          & ashcan                            & \checkmark                         \\
40           & cushion                           & \checkmark                         & 90           & television                        & \checkmark     & 140          & fan                               & \checkmark                         \\
41           & base                              & \checkmark                         & 91           & airplane                          & \checkmark     & 141          & pier                              &                                    \\
42           & box                               & \checkmark                         & 92           & dirt                              &                & 142          & crt                               & \checkmark                         \\
43           & column                            & \checkmark                         & 93           & apparel                           & \checkmark     & 143          & plate                             & \checkmark                         \\
44           & signboard                         & \checkmark                         & 94           & pole                              & \checkmark     & 144          & monitor                           & \checkmark                         \\
45           & chest                             & \checkmark                         & 95           & land                              &                & 145          & bulletin                          & \checkmark                         \\
46           & counter                           & \checkmark                         & 96           & bannister                         & \checkmark     & 146          & shower                            & \checkmark                         \\
47           & sand                              &                                    & 97           & escalator                         &                & 147          & radiator                          & \checkmark                         \\
48           & sink                              & \checkmark                         & 98           & ottoman                           & \checkmark     & 148          & glass                             & \checkmark                         \\
49           & skyscraper                        &                                    & 99           & bottle                            & \checkmark     & 149          & clock                             & \checkmark                         \\
50           & fireplace                         & \checkmark                         & 100          & buffet                            & \checkmark     & 150          & flag                              & \checkmark                        
\end{tabular}
\caption{Class ordering of ADE20K \cite{zhou2017adedataset} used in all reported experiments. }
\label{tab:cls_order}
\end{table*}
\section{Class Ordering}
In \cref{tab:cls_order} we report the class ordering of ADE20K that we used for all the reported experiments, following the previous benchmarks \cite{cermelli2020modelingthebackground, douillard2020plop}. Considering the \textit{100-50}, \textit{100-10}, and \textit{100-5} settings, reported in the main paper, we note that 44 of the new classes are “things", while the other 6 are “stuff". While there is no difference between “things" and “stuff" in semantic segmentation, it is especially relevant in the panoptic segmentation task, where the goal is to separate in different segments multiple instances of the “things" classes, since it introduces additional challenges.

\end{document}